\pdfoutput=1

\documentclass[11pt]{article}

\usepackage[preprint]{acl}

\usepackage{times}
\usepackage{latexsym}

\usepackage[T1]{fontenc}

\usepackage[utf8]{inputenc}

\usepackage{microtype}

\usepackage{inconsolata}

\usepackage{graphicx}

\usepackage{booktabs}
\usepackage{multirow}
\usepackage{colortbl}
\usepackage{amssymb}

\newcommand{\methodname}[0]{\textsc{CoRE}}
\newcommand{\methodnames}[0]{\textsc{CoRE}~}

\newcommand{\methodlongname}[0]{Combination of Retrieval Enrichment}

\title{Retrieval-enriched zero-shot image classification\\in low-resource domains}

\author{
 \textbf{Nicola Dall'Asen\textsuperscript{1,2}}
 \textbf{Yiming Wang\textsuperscript{3}}
 \textbf{Enrico Fini\textsuperscript{1}\thanks{Currently at Apple}}
 \textbf{Elisa Ricci\textsuperscript{1,3}}
\\
\\
 \textsuperscript{1}University of Trento
 \textsuperscript{2}University of Pisa
 \textsuperscript{3}Fondazione Bruno Kessler
\\
 \small{
   \textbf{Correspondence:} \href{mailto:nicola.dallasen@unitn.it}{nicola.dallasen@unitn.it}
 }
\\
\small{
   \textbf{Project website:} \href{https://fodark.github.io/CoRE}{https://fodark.github.io/CoRE}
 }
}

\usepackage{amsmath}
\usepackage{pifont}
\usepackage{svg}
\usepackage{adjustbox}
\usepackage{tabularx}
\usepackage{nicematrix}
\usepackage{makecell}

\newcommand{\PreserveBackslash}[1]{\let\temp=\#1\let\=\temp}
\newcolumntype{C}[1]{>{\PreserveBackslash\centering}p{#1}}

\definecolor{aqua}{rgb}{0.0, 1.0, 1.0}
\definecolor{ForestGreen}{RGB}{34,139,34}
\definecolor{RoyalBlue}{RGB}{204,255,94}

\definecolor{CoreCircuits}{RGB}{123,241,168}
\definecolor{CoreHam}{RGB}{254,200,154}
\definecolor{CoreInat}{RGB}{189,224,254}
\newcommand{\cmark}{\textcolor{ForestGreen}{\ding{51}}}
\newcommand{\xmark}{\textcolor{red}{\ding{55}}}

\newcommand{\eg}[0]{\textit{e.g.}~}
\newcommand{\ie}[0]{\textit{i.e.}~}
\newcommand{\wrt}[0]{\textit{w.r.t.}~}
\newcommand{\pp}[1]{\noindent\textbf{#1.}~}
\DeclareMathOperator*{\argmax}{arg\,max}

\DeclareMathOperator*{\topk}{top-k}

\newcommand{\captions}[0]{\mathcal{T}}
\newcommand{\clipfunc}[0]{{f}_{VLM}}
\newcommand{\clipimgenc}[0]{{f}_{VLM}^{V}}
\newcommand{\cliptxtenc}[0]{{f}_{VLM}^{L}}
\newcommand{\llmtxtenc}[0]{{f}_{LLM}}
\newcommand{\class}[0]{c}

\newcommand{\classes}[0]{\{c_n\}_{n=1}^{N}}
\newcommand{\imgemb}[0]{z_q}

\newcommand{\imagespace}[0]{\mathcal{V}}
\newcommand{\languagespace}[0]{\mathcal{L}}
\newcommand{\normalizedsimdef}[2]{\mathrm{softmax}\left(\frac{#1}{#2}\right)}

\newcommand{\textquery}[0]{p}
\newcommand{\textquerydef}[0]{\mathtt{``\{prefix\}~[CLS]_n"}}
\newcommand{\llmemb}[0]{l}
\newcommand{\zeroshotw}[0]{W}
\newcommand{\database}[0]{\mathbb{D}}

\newcommand{\txtcaptionssim}[0]{\mathcal{S}^T}
\newcommand{\txttxttemp}[0]{\tau_{t2t}}
\newcommand{\txtcaptionsemb}[0]{\mathcal{Z}^{T}}
\newcommand{\normalizedsimtxt}[0]{\sigma^{T}}
\newcommand{\zeroshotretrieved}[0]{W^T}
\newcommand{\zeroshotfinal}[0]{W^{+}}

\newcommand{\query}[0]{q}

\newcommand{\imgcaptionsembfinal}[0]{{z}^{T}}

\newcommand{\imgtxttemp}[0]{\tau_{i2t}}

\newcommand{\finalimgemb}[0]{z_q^{+}}

\begin{document}
\maketitle

\begin{abstract}
Low-resource domains, characterized by scarce data and annotations, present significant challenges for language and visual understanding tasks, with the latter much under-explored in the literature. Recent advancements in Vision-Language Models (VLM) have shown promising results in high-resource domains but fall short in low-resource concepts that are under-represented (\eg only a handful of images per category) in the pre-training set. 
We tackle the challenging task of zero-shot low-resource image classification from a novel perspective. By leveraging a retrieval-based strategy, we achieve this in a training-free fashion. 
Specifically, our method, named \methodnames(\methodlongname), enriches the representation of both query images and class prototypes by retrieving relevant textual information from large web-crawled databases. This retrieval-based enrichment significantly boosts classification performance by incorporating the broader contextual information relevant to the specific class. 
We validate our method on a newly established benchmark covering diverse low-resource domains, including medical imaging, rare plants, and circuits. Our experiments demonstrate that \methodname~outperforms existing state-of-the-art methods that rely on synthetic data generation and model fine-tuning. 
\end{abstract}

\section{Introduction}
\label{sec:intro}

Low-resource domains refer to those rare domains where the data or its annotation is truly scarce. Similarly, low-resource languages are those that have significantly less content available online~\cite{magueresse2020low} with respect to other high-resource languages, like English. There exist abundant research on the topic in the context of natural language processing~\cite{ranathunga2023neural,adams2017cross,fadaee2017data,pan2017cross}.
However, surprisingly, the vision counterpart, \ie \textit{low-resource visual domains}, is much under-explored despite the numerous practical applications. In this paper, we focus on classifying images in low resource domains, \ie where we can find only a handful of images per category. 
The causes for such limited data can be various: for example, when only certain devices are capable of capturing the visual content, \eg astronomy or medical imaging; the visual content itself is sensitive or private, \eg due to privacy issues, or rarely appears in nature, \eg deep ocean animals, or other long-tailed categories. Their associated annotations can also be limited due to the expertise requires, in particular for niche fields, \eg electric design or phytology.

\begin{figure}[t!]
    \centering
    \includegraphics[width=1.0\linewidth]{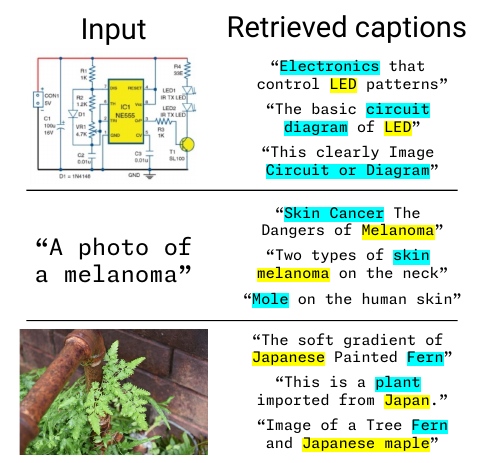}
    \caption{Our retrieval-based solution enriches both images and textual descriptors with real-world captions which contain \colorbox{aqua}{domains} and \colorbox{yellow}{classes}. Even when the captions are generic (third row for each example), they can still restrict the focus to the correct domain.}
    \vspace{-0.5cm}
    \label{fig:teaser}
\end{figure}

Recent large vision-language models (VLMs) have fostered a paradigm shift in image classification. Their flexibility and generalization, enabled by web-scale pre-training with text-image pairs, makes them versatile tools in many sub-fields of computer vision. Numerous works have appeared, with the objective of tuning VLMs, \eg CLIP~\cite{radford2021learning} or SigLIP~\cite{zhai2023sigmoid}, to address image  zero-shot~\cite{jia2022visual} or few-shot~\cite{chowdhury-etal-2023-apollo, da2023diversified} image classification.
However, the images involved in those studies are mostly in \textit{high-resource} image domains, where there exist thousands of images on the Internet for VLMs to learn from during the web-scale pre-training~\cite{udandarao2024no}. 

However, directly performing zero-shot classification in low-resource domains does not yield satisfactory performance due to the data scarcity in pre-training. Even supervised fine-tuning might fall short in learning the underlying data distribution due to the very limited amount of data and annotation. 
Among the techniques that have been explored in the pioneering work~\cite{zhang2024low}, one prominent recipe is to fine-tune the VLMs on data augmented via synthetic generation (\eg Stable Diffusion~\citep{rombach2021high}). 
Despite the performance improvements, by analyzing the generated images, we observe that image generation models are also affected by the low-resource nature of the task. The generation quality is largely dependent on the noise injected on the real samples: by injecting limited noise, the synthetic images appear very similar to the original samples, being correct but not diverse, while by injecting more noise, the synthetic images diversify in appearance, but are mostly semantically incorrect and exhibit domain-specific rule violations. This is because the data distribution of rare domains is not well-represented in the generative models latent space~\citep{mokady2022null,trabucco2024effective}.

Instead of generating  synthetic images as data augmentation, we explore the possibility of retrieving relevant information from a textual corpus, crawled from the Internet, to enrich the data representation at inference time, as shown in Fig.~\ref{fig:teaser}. 
It turns out that retrieval is also non-trivial in the low-resource regime as (i) pre-trained models generally under-represent the low-resource domains, thus greatly limiting the retrieval efficacy; (ii) large web-crawled databases can contain noisy or incorrect content, a problem that is more severe in low-resource domains.

Thus, a careful design is required to leverage the retrieved data. In this work, we propose the first training-free and retrieval-based method, \methodname~(\methodlongname), to tackle low-resource image classification.
Following a VLM-based zero-shot classification paradigm, we propose to enrich the representation for both the query image and the class prototypes with textual content retrieved with different encoder backbones from large web-crawled databases. 
Specifically, for the query image, we employ the pre-trained image encoder from a VLM as our vision retrieval backbone. We perform image-to-text retrieval, obtaining the most relevant captions with respect to the query image.

From our preliminary analysis, we observe that although the specific category (\eg ``LED'') appears sparsely in the retrieval, its broader category (``circuit'') does occur frequently. Previous studies have empirically demonstrated that enriching the prompt with the broader concept, together with noise, can significantly boost the zero-shot recognition performance~\cite{roth2023waffling}. 
We thus enrich the image embedding by combining it with the textual embedding from image-to-text retrieval. 
Similarly, we construct the enriched class prototypes. For each class, we form its corresponding text prompt and embed it with a pre-trained text encoder to retrieve captions that are most relevant. Then, the retrieved captions are encoded with the VLM text encoder and aggregated together with the textual embedding of the original class prompt. The final categorization is obtained by computing the cosine similarity between the enriched visual representation against the enriched textual class prototypes.

To validate the effectiveness of our proposed method, we also collect a set of datasets that covers diverse low-resource domains, including medical imaging, rare plants, and circuits.
\methodname~can effectively improve the data representation in a training-free fashion, with a noticeable improvement in image classification performance on all the datasets, outperforming the state-of-the-art method that involves synthetic image generation and model fine-tuning. 

To summarize, our contributions are:
\begin{itemize}
    \item We propose the first training-free retrieval-based method \methodname~for addressing zero-shot low-resource image classification;

    \item we propose a data representation enrichment strategy for both query image and class prototypes, using the textual content retrieved from the database;

    \item we establish a benchmark featuring zero-shot low-resource image classification, composed of representative datasets and VLM-based baselines and state-of-the-art methods;

    \item our training-free method is effective in classifying low-resource images, outperforming competitors with training and other training-free baselines by a large margin.
\end{itemize}

\section{Related work}
\label{sec:sota}

\pp{High-resource data}The \textit{de-facto} standard in Deep Learning has become to train larger foundation models with a high volume of data, \eg ImageNet~\citep{imagenet} or LAION-5B~\citep{schuhmann2022laion} for vision, which contain up to 5 billion images, or FineWeb~\citep{penedo2024fineweb} for text, which contains 15 trillion of tokens. For vision-related tasks, data focuses on natural images, which are plentiful online and can be easily obtained. However, when moving to more specific domains, \eg medical~\cite{irvin2019chexpert}, satellite imaging~\cite{helber2019eurosat}, or long-tailed distributed data~\cite{van2017inaturalist}, data becomes less available. Although these rare domains have been understudied in favor of higher-available data, we follow the study of~\citep{zhang2024low} and investigate domains where the number of available data is in the order of hundreds, and training is an under-performing option.

We propose a novel way to address the challenge, and we exploit the knowledge coming from web-scale image-text pairs datasets through retrieval. Retrieval-based solutions have proved successful for image classification~\citep{liu2023learning,conti2023vocabularyfree} and NLP tasks~\citep{lewis2020retrieval}, and we propose to enrich the data representation in VLMs.

\pp{Multimodal foundation models} Multimodal Foundation Models, such as CLIP~\citep{radford2021learning} or BLIP~\citep{li2022blip,li2023blip} have gained a lot of popularity due to their outstanding zero-shot capabilities, derived from their weak-supervised training on web-crawled data. The most common paradigm is to train on image-text pairs, which can be easily obtained on the web~\cite{changpinyo2021cc12m,schuhmann2022laion}, but recent approaches like ImageBind~\citep{girdhar2023imagebind} bridge several modalities, \eg images, videos, audio, text, and thermal. We focus in particular on Vision-Language Models (VLM).

Adaptation of VLM to downstream tasks is performed in several ways, \eg full finetuning, or Parameter-Efficient FineTuning (PEFT) methods, \eg LoRA~\citep{hu2022lora}, Prompt Tuning~\citep{jia2022visual}, or Textual Prompt Tuning~\citep{zhou2022learning}. These approaches are not suitable for the rare domain setting as the amount of data is limited, domain-specific, and extremely different from the pre-training data of the original VLM.

\pp{VLM few-shot learning} Several works have proved the effectiveness of few-shot learning when adapting VLMs to downstream tasks. Some notable works include a combination of PEFT strategy with VLMs, such as APoLLo~\citep{chowdhury-etal-2023-apollo} that synthetically augments both the visual and textual branch of CLIP, or DISEF~\citep{da2023diversified} that employs LoRA and synthetic images to fine-tune CLIP. We closely follow the work of~\citep{zhang2024low}, where the authors fine-tune ImageBind~\citep{girdhar2023imagebind} with an AdaptFormer~\citep{chen2022adaptformer} module on a combination of real and synthetic data for low-resource rare domains. Differently from these works, we do not employ synthetic data and we do not fine-tune, instead, we propose a training-free zero-shot solution through retrieval to adapt both the visual and textual representation in CLIP.

\section{Background}

\begin{figure*}[t!]
    \centering
  \includegraphics[width=1.0\linewidth]{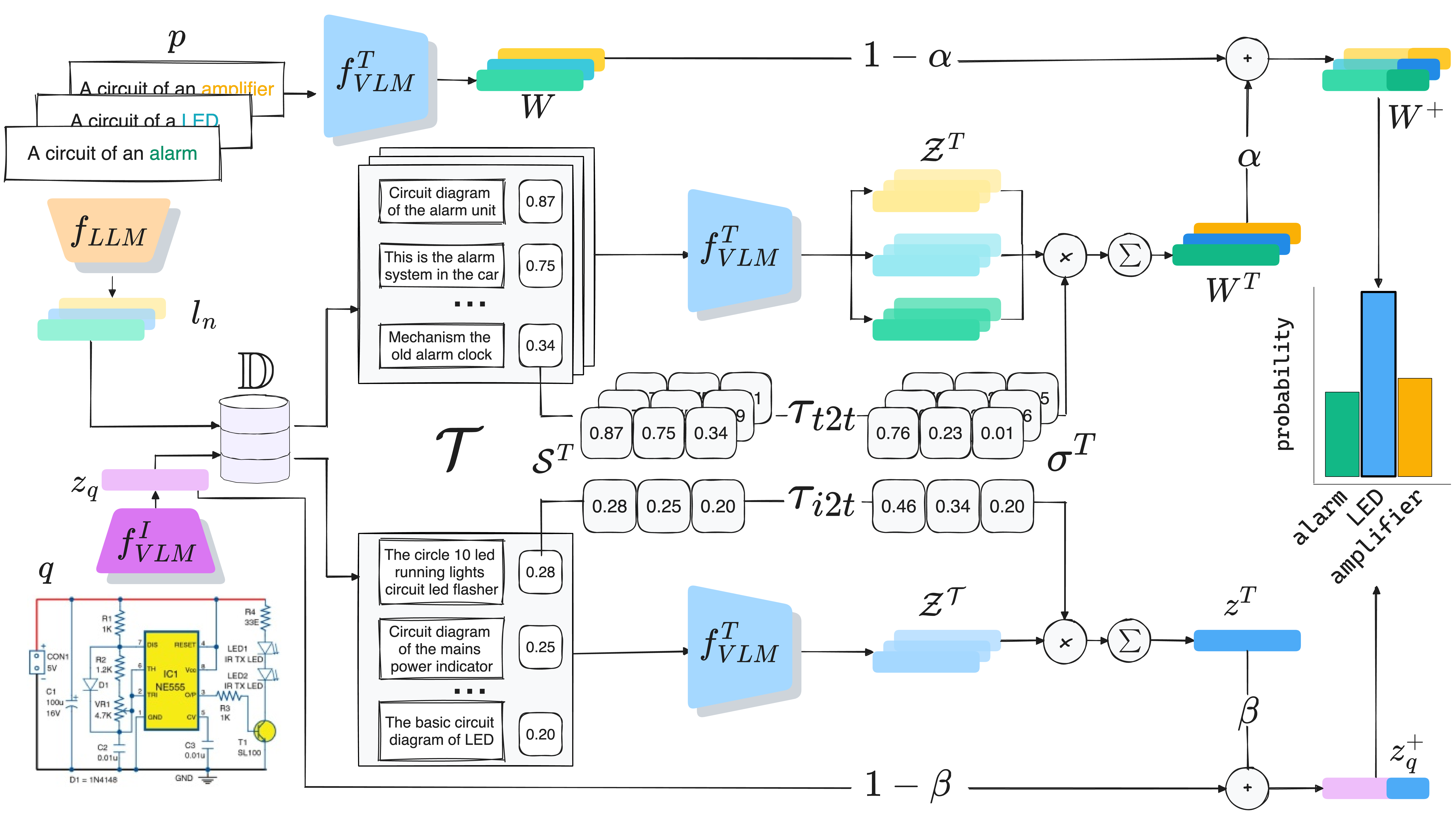}
  \caption {Our \methodnames enriches both the image embedding $z_q$ and the class prompts $p$ with retrieved captions from a large-scale web-crawled database $\database$. We weight the retrieved captions $\mathcal{T}$ with their similarity scores $\txtcaptionssim$, which we skew with controllable temperatures $\imgtxttemp$ and $\txttxttemp$. By combining the retrieved captions embedding with the original representations $W$ and $q$ through $\alpha$ and $\beta$, we obtain enriched representations $W^+$ and $z_q^+$ which we employ for zero-shot classification.}
  \label{fig:method}
\end{figure*}

To establish a foundation in understanding our method, we provide a brief introduction to Vision-Language Models and web-scale databases, the two essential elements in our method design.
\subsection{Vision-Language Models}\label{sec:vlm}
Vision-Language Models (VLM)~\citep{radford2021learning,jia2021scaling} learn a function $\clipfunc: \imagespace\times\languagespace\rightarrow\mathbb{R}$ with the goal to maximize their representation similarities $\mathbb{R}$ in order to map images in visual space $\imagespace$ and texts in language space $\languagespace$ to the same latent space. 
In particular, a VLM is composed of a vision encoder $\clipimgenc: \imagespace\rightarrow\mathbb{R}^{d}$ that maps images to a visual embedding and a language encoder $\cliptxtenc: \languagespace\rightarrow\mathbb{R}^{d}$ that maps the text in natural language (after converting into tokens) to an embedding. VLMs learn to project the two modalities to the same latent space $\mathbb{R}^{d}$ via contrastive learning with millions of web-crawled image-text pairs, enabling image classification using text by evaluating their similarity in this shared space. 

At inference, the VLM can be used to assess the similarity between an image sample and a set of prompt texts that are composed from a number of $N$ classes $\classes$~\cite{radford2021learning}. 
Given a query image $\query \in \imagespace$, we can obtain its visual embedding via the image encoder as $\imgemb = \clipimgenc(\query)$.
We can build the class prototypes $\zeroshotw \in \mathbb{R}^{N\times d}$ as the textual embeddings using the VLM text encoder. Specifically, for each class $c_n$, we build the text prompt $\textquery_n = \textquerydef$, where \texttt{\{prefix\}} usually is \texttt{``a photo of a''} and $\mathtt{[CLS]}_n$ is the name of $c_n$ in text. Then the class prototypes can be formed as $\zeroshotw = \cliptxtenc(\{\textquery\}_n)$.
Finally, we can compute the cosine similarities between the image embedding $\imgemb$ and the class prototypes $\zeroshotw$ in order to predict the class $\hat{c}$ for the query image $\query$:
\begin{equation}
    \label{eq:zeroshot}
    \hat{c} = \argmax_c (\imgemb \times \zeroshotw^{\intercal}).
\end{equation}

\subsection{Retrieval databases}

Together with the VLMs, there is also the emergence of large vision-language databases that can be leveraged for VLM pre-training and retrieval. The community has collected and released several web-scale image-text pairs datasets, such as LAION~\cite{schuhmann2022laion}, CC12M~\cite{changpinyo2021cc12m}, and COYO~\cite{kakaobrain2022coyo-700m}, with respectively 400M/5B, 12M and 700M image-text pairs. Formally, let $\database = \{(i,t)_m\}_{m=1}^{M}$ be the retrieval database that consists of $M$ items, where each item is paired with an image $i_m$ and its associated textual description $t_m$. We are mostly interested in the textual content of $\database$ as it contains rich language-induced semantics~\cite{conti2023vocabularyfree} that might contributes to enriching the knowledge that is specific low-resource domain. 

With millions and billions of data items, it is critical performing retrieval efficiently. To this purpose, we focus on embedding-based retrieval, whose goal is to retrieve the most similar elements from a database from a query embedding. Such retrieval is supported by off-the-shelf tools, \eg Faiss~\cite{douze2024faiss}, that enable similarity search and clustering of dense vectors at scale.
In particular, we prepare the text-image database $\database$ by encoding data items into embeddings using encoders of VLMs. Given an embedding $z$ corresponds to either visual or textual modality, we can retrieve a set of $k$ textual descriptions $\captions$ that are most similar to the $z$ from the database $\database$:
\begin{equation}
    \captions = \topk_t (\langle z, \cliptxtenc(t)\rangle), \forall t \in \mathbb{D},
\end{equation}
where $\langle \cdot \rangle$ computes the cosine similarity between two embeddings, and $\topk (\cdot)$ returns $k$ textual descriptions with the highest cosine similarities.

\section{\methodname}
\label{sec:method}

We focus on zero-shot low-resource image classification, following the paradigm of VLMs as described in Sec~\ref{sec:vlm}. Low-resource domains are generally not well represented by VLMs given their limited availability in the pre-training dataset~\cite{udandarao2024no}. There is a need to steer the original data representation towards more specialized low-resource domains represented by the query image $\query$. 

To this end, our training-free method \methodname~extracts domain-relevant information from a large text-image database $\database$, to enrich the representation of both class prototypes and the query image. 

For the class prototypes, \ie textual representation of the classes $\classes$, we retrieve semantically close captions for each class with the embedding of prompt text regarding the $\class$. 
On one hand, the retrieved captions contain rich domain-level information while being less specific to the exact class. On the other hand, the class prompt is only specific to the class of interest without much prior of the domain information. Therefore, we further join the retrieved captions with the prompt class text to obtain the textual class prototypes $\zeroshotfinal$ enriched with the domain context. 

A similar rationale is applied to enrich the query image representation via image-to-text retrieval, implemented by the image encoder $\clipimgenc$ of a pre-trained VLM. With the visual embedding $\imgemb$ of the query image $\query$, we retrieve the set of captions that are the most aligned to $\imgemb$ in the shared latent space, and use the retrieved captions to obtain an enriched image representation $\finalimgemb$.
The final class is predicted using $\finalimgemb$ and $\zeroshotfinal$ similar to Eq.~\ref{eq:zeroshot}: 
\begin{equation}
    \label{eq:final}
    \hat{c} = \argmax_c (\finalimgemb \times {\zeroshotfinal}^{\intercal}).
\end{equation}
We describe each retrieval branch in detail in the following sections and we show an overview of our \methodnames in Fig.~\ref{fig:method}.

\subsection{Class representation enrichment}

We leverage text-to-text retrieval to enrich the class prototypes of the set of predefined classes $\classes$. 
For each class $c_n$, we first build the text prompt $\textquery_n$ following the prompt format as described in Sec.~\ref{sec:vlm}. Different prompt templates are also experimented in Sec.~\ref{sec:ablation}. 

For the text-to-text retrieval, we leverage the encoder of a LLM to obtain the textual embeddings for both the per-class text prompt, \ie $\llmemb_n = \llmtxtenc(\textquery_n)$ and all textual content in the database $\database$.
For each $\llmemb_n$, we then retrieve from $\database$ a set of $k$ most similar textual descriptions $\captions_n$ with respect to class $c_n$.
Each retrieved text $t_i\in \captions_n$ has an associated cosine similarity in the range of $[-1,1]$ \wrt the embedding of the prompt text $\llmemb_n$, forming a set of $k$ scores $\txtcaptionssim_n$.

As the retrieved texts contain rich domain-level context, we further embed them and merge their textual embeddings to obtain a domain-specialized embedding. Since the eventual classification is achieved in the latent space of VLMs, we leverage the text encoder of the VLM to embed the retrieved texts, \ie $\txtcaptionsemb_n = \cliptxtenc(\captions_n)$. 
As the retrieved captions are associated with different similarity scores $\txtcaptionssim_n$, we weigh their contribution to form the domain-specialized embedding accordingly. Specifically, we propose to build a probability distribution out of the similarities scores as:
\begin{equation}
    \normalizedsimtxt_n=\normalizedsimdef{\txtcaptionssim}{\txttxttemp},
\end{equation}
where $\txttxttemp$ is the temperature parameter that controls the skewness of the distribution.

For the class $c_n$, the embedding from retrieved texts $\txtcaptionsemb_n$ are then combined as a weighted sum with the weight being $\normalizedsimtxt_n$, forming the domain-specialized embeddings for all classes $\zeroshotretrieved$.

Finally, we build the retrieval-enriched class prototypes by linearly interpolating prompt-text class prototypes $\zeroshotw$ as described in Sec~\ref{sec:vlm} and the retrieved domain-specialized ones $\zeroshotretrieved$, with a interpolation factor $\alpha$ as:
\vspace{-0.3cm}
\begin{equation}
    \zeroshotfinal = \alpha\zeroshotretrieved + (1-\alpha)\zeroshotw,
\end{equation}
where $\alpha$ is a hyperparameter and we shows its impact in Section~\ref{sec:ablation}.

\begin{table*}[!th]
\centering
\resizebox{\linewidth}{!}{
\begin{tabular}{ccccccc}
\toprule
\multirow{2}{*}{\textbf{Method}}              & \multicolumn{2}{c}{\cellcolor{CoreCircuits}\textbf{Circuits}}       &  \multicolumn{2}{c}{\cellcolor{CoreInat}\textbf{iNaturalist2021 (LT100)}}  & \multicolumn{2}{c}{\cellcolor{CoreHam}\textbf{HAM10000}}       \\ 
& \cellcolor{CoreCircuits}\textbf{Acc@1} & \cellcolor{CoreCircuits}\textbf{Acc@5} & \cellcolor{CoreInat}\textbf{Acc@1} & \cellcolor{CoreInat}\textbf{Acc@5} & \cellcolor{CoreHam}\textbf{Acc@1} & \cellcolor{CoreHam}\textbf{Acc@5} \\
\midrule
ImageBind \raisebox{-1.mm}{\includegraphics[height=4.0mm]{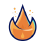}}~\cite{zhang2024low} & 24.10 & 49.30 & 31.60$^\dagger$ & 60.50$^\dagger$ & 54.60$^\dagger$    & 96.56$^\dagger$     \\
SigLIP@384px \raisebox{-1.mm}{\includegraphics[height=4.0mm]{images/icon_fire.pdf}}~\cite{zhai2023sigmoid}    & 19.53$^\dagger$   & 30.61$^\dagger$       & 34.50$^\dagger$ & 63.50$^\dagger$ & 54.60$^\dagger$       & 95.90$^\dagger$      \\
\midrule
CLIP ViT-L~\cite{radford2021learning}           & 7.98 & 29.13          &  8.00 & 22.60  & 45.27 & 90.80  \\
CLIP ViT-L@336px~\cite{radford2021learning}           & 9.09   & 30.33        & 7.60 & 22.70 & 40.97 & 90.27  \\
BLIP2-EVA~\cite{li2023blip}           & 17.63   & N/A       &  1.40 & N/A & 2.91 & N/A  \\
LlaVA 1.6 34B~\cite{liu2023improved} & 29.59 & N/A & 0.60 & N/A & 10.59 & N/A \\
ImageBind~\cite{girdhar2023imagebind}           & 22.36    & 51.02      &  6.70 & 23.90 & 14.43  & 84.25 \\
SigLIP@384px~\cite{zhai2023sigmoid}              & 35.81    & 58.63      & 19.10 & \textbf{45.70} & 57.64 & \textbf{96.16} \\
\methodnames (Ours --- CC12M)                & $\mathrm{\underline{42.94}}_{\mathcolor{ForestGreen}{~7.13}}$ & $\mathrm{\underline{67.71}}_{\mathcolor{ForestGreen}{~9.08}}$ & $\mathrm{\underline{21.40}}_{\mathcolor{ForestGreen}{~2.30}}$ & $\mathrm{42.59}_{\mathcolor{red}{~3.11}}$ & $\mathrm{\underline{61.54}}_{\mathcolor{ForestGreen}{~3.90}}$    & $\mathrm{\underline{95.70}}_{\mathcolor{red}{~0.46}}$      \\

\methodnames (Ours --- COYO-700M) & $\mathrm{\textbf{43.88}}_{\mathcolor{ForestGreen}{~8.07}}$ & $\mathrm{\textbf{71.99}}_{\mathcolor{ForestGreen}{~13.36}}$ & $\textbf{22.10}_{\mathcolor{ForestGreen}{~3.00}}$ & $\underline{44.10}_{\mathcolor{red}{~1.60}}$ & $\mathrm{\textbf{62.21}}_{\mathcolor{ForestGreen}{~4.57}}$ & $94.51_{\mathcolor{red}{~1.65}}$ \\ 
 \bottomrule
\end{tabular}
}
\caption{Top-1 and top-5 accuracy on the proposed benchmark. $^\dagger$ indicates our re-implementation. \raisebox{-1.mm}{\includegraphics[height=4.0mm]{images/icon_fire.pdf}} denotes fine-tuning. We highlight \textbf{best} and \underline{second best} results.
We report \textcolor{ForestGreen}{gain} and \textcolor{red}{loss} \wrt the best training-free solution.}
\label{tab:quantitative}
\end{table*}

\subsection{Image query representation enrichment}
Symmetrically, we want to enrich the query image representation and exploit the rich web-scale semantics of captions to build a more representative image embedding. 

Starting from a query image $\query$. For image-to-text retrieval, we use the visual encoder of the VLM to obtain the image embedding $\imgemb = \clipimgenc(\query)$, while we use the text encoder $\cliptxtenc$ to embed the database $\database$, in this way image and text are mapped in the same space. We retrieve from $\database$ the $k$ most similar captions $\captions$, with their associated cosine similarities $\txtcaptionssim$. Similar to the class prototype branch, we use the scores to build a probability distribution:

\begin{equation}
\normalizedsimtxt=\normalizedsimdef{\txtcaptionssim}{\imgtxttemp},
\end{equation}

where $\imgtxttemp$ controls the final distribution skewness.

We then embed the retrieved captions with the VLM text encoder $\txtcaptionsemb=\cliptxtenc(\captions)$, and combine them as a weighted sum using $\normalizedsimtxt$, obtaining $\imgcaptionsembfinal$.

Finally, the original query image embedding $\imgemb$ and the retrieved captions embedding $\imgcaptionsembfinal$ are linearly interpolated to build the final query representation as:
\begin{equation}
    \finalimgemb = \beta \imgcaptionsembfinal + (1-\beta) \imgemb,
\end{equation}
where the hyperparameter $\beta$ controls the importance of the two components and we study the effect of this parameter in Section~\ref{sec:ablation}.

\section{Experiments}

We evaluate our proposed method \methodnames in comparison with state-of-the-art Vision-Language approaches using three challenging low-resource datasets. We describe the dataset used and the evaluation protocol we follow. We discuss the results \wrt the baseline methods, and we ablate the proposed components and architectural choices.

\pp{Datasets} We consider three datasets covering different low-resource scenarios for our analysis: we employ the Circuit Diagram Classification dataset~\cite{zhang2024low}\footnote{The other released datasets relate to retrieval tasks.} that comprises 1,332 circuit diagram images covering 32 different classes, the images are scraped from the web and textbooks. The authors split the data into 154 images for training and the remaining for testing, resulting in an average of $\sim5$ samples per class available at training time.

The second dataset we consider is iNaturalist 2021~\citep{van2017inaturalist} which contains 10,000 species with a fine-grained classification. The dataset features many visually similar species, captured in a wide variety of situations. In order to remain in the rare domain setting, we restrict our analysis to the rarest 100 species in terms of available training samples and set the maximum amount of training shots to 5. We test on 10 images for each class, therefore the test set contains 1,000 samples. 

The last dataset we employ is HAM10000~\citep{tschandl2018ham10000}, which comprises dermatoscopic images from different populations. The dataset has 7 classes that include a representative collection of all important diagnostic categories in the realm of pigmented lesions. The test set includes 1,511 images.

\pp{Database(s)} We use CC12M~\cite{changpinyo2021cc12m} as our source dataset to build the database $\database$, following previous works~\cite{conti2023vocabularyfree}. We also use a subset (10\%) of COYO-700M to test the scaling laws of the retrieval database. We focus on their textual corpus as we are interested in retrieving the relevant pieces of text. The retrieval databases are implemented using Faiss~\citep{johnson2019billion} on pre-extracted embeddings. We use the SigLIP~\citep{zhai2023sigmoid} text encoder for the image-to-text retrieval, while we employ a text-only encoder to obtain a stronger textual semantic in text-to-text retrieval. In particular, we use SFR-Embedding-Mistral~\citep{SFRAIResearch2024} as $\llmtxtenc$.

\pp{Evaluation protocol} As a common practice in image classification tasks, we report the performance as the top-1 and top-5 accuracy. For VLMs, we compare the index of the highest logit to the ground truth, while for Large Multimodal Models (LMM) we parse the LMM output to extract the predicted class name and match it with the dataset class names.

\pp{Implementation details} 
\methodnames leverages an LLM encoder $\llmtxtenc$, implemented as SFR-Embedding-Mistral~\cite{SFRAIResearch2024}, an image encoder $\clipimgenc$, implemented as the SigLIP~\cite{zhai2023sigmoid} vision encoder, and a text encoder $\cliptxtenc$, implemented as the SigLIP text encoder. 
The number $k$ of retrieved captions is set to $10$ following previous works~\cite{conti2023vocabularyfree}, nevertheless, the temperature parameters $\txttxttemp$ and $\imgtxttemp$ allow restricting the effect of the lower-ranked retrieved captions by skewing the distribution towards the most similar samples.

For our \methodname, we empirically find that a high temperature $\imgtxttemp$ for image-retrieved captions leads to more favorable results, therefore we set it to $100$. For text-retrieved captions, a lower temperature is more beneficial and we set it to $1$, skewing the distribution more towards the high-confidence samples. For the effect of $\alpha$ and $\beta$ we refer to Sec.~\ref{sec:ablation}. 

For the zero-shot and retrieval prompting strategies, we find that having a domain-specific prompt for zero-shot and a generic prompt for retrieval leads to favorably good results across all the datasets, and we provide a complete study of this in Sec.~\ref{sec:ablation}. For implementation details of the baselines we refer to Appendix~\ref{app:implementation}

\subsection{Comparisons}
\pp{Baselines} We compare against state-of-the-art VLMs and LMMs in the training-free zero-shot scenario. In particular, we compare with CLIP~\citep{radford2021learning}, using the ViT-L/14 and ViT-L/14@336px vision encoder variants, BLIP-2~\citep{li2023blip} using the ViT-g/14 vision encoder from EVA-CLIP~\citep{fang2023eva}, LLaVA-NeXT~\citep{liu2023improved} with a 34B LLM, ImageBind~\citep{girdhar2023imagebind} as in~\citep{zhang2024low}, and SigLIP~\citep{zhai2023sigmoid}. For the last two models, we also implement a training-based variant as in~\citep{zhang2024low}, which trains an AdaptFormer~\citep{chen2022adaptformer} and a linear classification head. 

\pp{Discussion}We present a quantitative evaluation of \methodnames and the baselines in Tab.~\ref{tab:quantitative}. We can see that among the training-free approaches, \methodnames outperforms the others by a substantial margin (up to $8.07\%$ in top-1 accuracy on Circuits). When comparing to fine-tuned solutions (\raisebox{-1.mm}{\includegraphics[height=4.0mm]{images/icon_fire.pdf}}), \methodnames still outperforms them, even given the substantial gain between zero-shot and fine-tuned ImageBind (up to $40\%$ in HAM10000). 

Fine-tuning SigLIP seems detrimental to performance, except for iNaturalist, and we deem the lower performance \wrt fine-tuned ImageBind to the higher number of trainable parameters, as SigLIP embeddings are bigger than ImageBind ones (1152 vs 1024). We study this behavior further in Section~\ref{sec:ablation}. 

LMMs reach satisfactory results on Circuits, while the low performance on iNaturalist and HAM10000 is due to these models answering consistently with the same class name for almost all the samples.

On iNaturalist, supervised fine-tuning represents a stronger solution \wrt our \methodname, and we deem this results to two factors: the nature of the dataset and the availability of the concepts in the retrieval database. Being the images of different plants and animals, the visual overlap is less prominent than in the other two datasets, therefore supervised fine-tuning can effectively separate different classes with a small amount of training data. Secondly, as in iNaturalist class names are provided with their Greek or Latin scientific name, this limits the amount of relevant data that can be retrieved from the database as even in large-scale image-text datasets, such as LAION 400M, this type of data is under-represented~\cite{parashar2023prompting}. 
We mitigate this effect by using both scientific and common names when retrieving and building the zero-shot weights. 

This benchmark allows us to show the training-free strength of our method, but also showcase the viability of training-based solutions given the right assumptions. Nevertheless, \methodnames remains the strongest training-free solution. We can also observe that employing a bigger retrieval database improves the top-1 accuracy across all the datasets.

\subsection{Ablation}
\label{sec:ablation}

\begin{table}[!t]
\centering
\resizebox{\linewidth}{!}{
\begin{tabular}{cccccccccc}
\toprule

\multirow{2}{*}{$\alpha$} & \multirow{2}{*}{$\txttxttemp$} & \multirow{2}{*}{$\beta$} & \multirow{2}{*}{$\imgtxttemp$} & \multicolumn{2}{c}{\cellcolor{CoreCircuits}\textbf{Circuits}}  & \multicolumn{2}{c}{\cellcolor{CoreInat}\textbf{iNaturalist}} & \multicolumn{2}{c}{\cellcolor{CoreHam}\textbf{HAM10000}} \\ 
& & & & \cellcolor{CoreCircuits}\textbf{Acc@1} & \cellcolor{CoreCircuits}\textbf{Acc@5} & \cellcolor{CoreInat}\textbf{Acc@1} & \cellcolor{CoreInat}\textbf{Acc@5} & \cellcolor{CoreHam}\textbf{Acc@1} & \cellcolor{CoreHam}\textbf{Acc@5} \\
\midrule
    \xmark  &   \xmark   &   \xmark   &  \xmark    &    35.81 & 58.63         &  19.10&\textbf{45.70} &  57.64&\textbf{96.16}   \\
   \cmark   &   \xmark   &   \xmark   &  \xmark    &    36.46&64.01        & 20.90&43.99 &   60.75&95.76    \\
   \cmark   &  \cmark    &   \xmark   &  \xmark    &    37.66&65.88         &  21.09&43.90  &  61.02&95.57  \\
   \cmark   &   \cmark   &   \cmark   &  \xmark    &    42.39&\textbf{69.85}         &   20.80&42.39  &  \textbf{61.55}&95.43 \\
   \cmark   &  \cmark    &   \cmark   &  \cmark    &     \textbf{42.94}&67.71        &   \textbf{21.40}&42.59 &  \textbf{61.55}&95.70  \\ \bottomrule

\end{tabular}
}
\caption{Ablation of proposed components of \methodnames CC12M, starting from the zero-shot only to the full \methodnames. As shown in the table, each of the components can bring performance improvement across all the datasets, proving the effectiveness of the designed retrieval strategy.}
\vspace{-0.1cm}
\label{tab:comp_abl}
\end{table}

\begin{figure}[!t]
    \centering
  \includegraphics[width=1\linewidth]{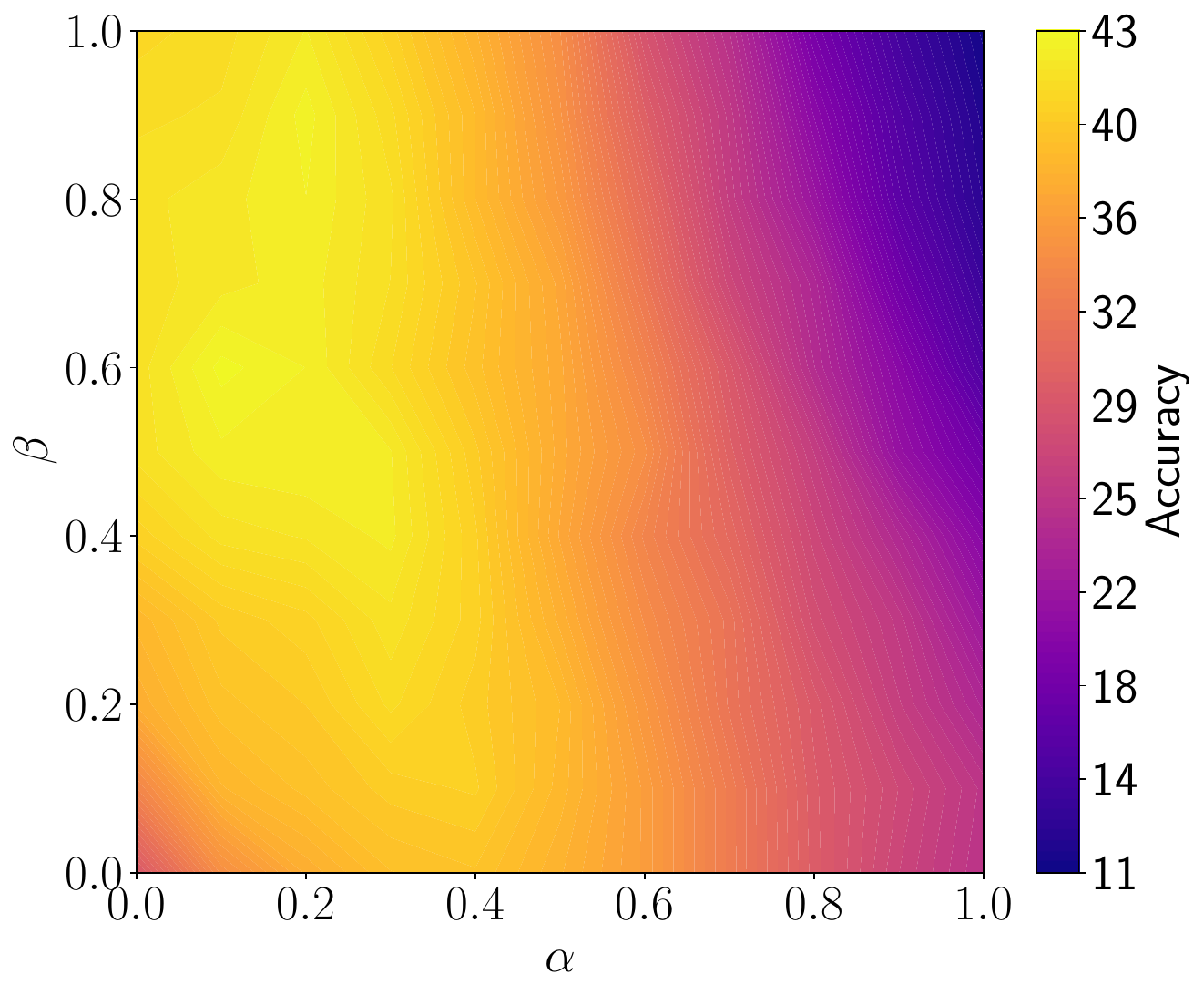}
  \caption {Top-1 accuracy of \methodnames CC12M on Circuits with varying $\alpha$ and $\beta$. \methodnames achieves the best performance with a balanced merge of image-retrieved captions ($\beta\sim 0.5$), while for class-relevant captions the best weighting is slightly lower ($\alpha\sim 0.2$).}
  \vspace{-0.3cm}
  \label{fig:alpha_beta}
\end{figure}

\pp{Various embedding fusion strategies} We ablate the contribution of each of our proposed components in Tab.~\ref{tab:comp_abl}, starting from the zero-shot only and reaching the full \methodname. We first add the weighting between zero-shot weights and retrieved weights without the temperature on the similarities, effectively having a naïve average of retrieved embeddings. We then introduce $\txttxttemp$ to skew the distribution. We thus move to the image retrieval part, where we first add a naïve average embedding weighted by $\beta$, and finally the weighting on the similarities $\imgtxttemp$. We can see how each of the components brings an improvement in performance across all the datasets, proving the effectiveness of our retrieval strategies.

We further isolate the effect of the two merging weights $\alpha$ and $\beta$ on the Circuits dataset in Fig.~\ref{fig:alpha_beta}, where we can see that \methodnames achieves the best performance with a balanced merge of image-retrieved captions ($\beta\sim 0.5$), while for class-relevant captions the best weighting is slightly lower ($\alpha\sim 0.2$). This indicates that the branch that obtains the most improvement from retrieval is the image one, while the zero-shot weights obtained from the classes are already representative, but they can still benefit from retrieved captions.

\pp{On the use of LLM for embedding} We ablate the choice of using an LLM to encode the text for the textual branch and compare this choice to using the VLM textual encoder $\cliptxtenc$. In particular, we use the SigLIP@384px text encoder and perform both image-to-text and text-to-text retrieval in the SigLIP latent space. 
From Tab.~\ref{tab:llm_ablation}, we see that LLM embeddings are stronger in the unimodal text-to-text retrieval. While SigLIP embedding is less performing than the LLM embeddings at text-to-text retrieval, the final classification performance is still better than all the other training-free baselines. 
Therefore SigLIP text embeddings remain a viable solution in case deploying the LLM (7B parameters) is infeasible due to computational constraints.

\begin{table}[!t]
    \centering
    \resizebox{\linewidth}{!}{
    \begin{tabular}{ccccccc}
    \toprule
       \multirow{2}{*}{\textbf{Encoder}}  & \multicolumn{2}{c}{\cellcolor{CoreCircuits}\textbf{Circuits}} & \multicolumn{2}{c}{\cellcolor{CoreInat}\textbf{iNaturalist}} & \multicolumn{2}{c}{\cellcolor{CoreHam}\textbf{HAM10000}}  \\ 
       & \cellcolor{CoreCircuits}\textbf{Acc@1} & \cellcolor{CoreCircuits}\textbf{Acc@5} & \cellcolor{CoreInat}\textbf{Acc@1} & \cellcolor{CoreInat}\textbf{Acc@5} & \cellcolor{CoreHam}\textbf{Acc@1} & \cellcolor{CoreHam}\textbf{Acc@5} \\ \midrule
       SigLIP%
       & 41.46 & 64.56 & 20.90 & \textbf{43.20} & 60.02 & 90.53 \\ 
       Mistral%
       & \textbf{42.94} & \textbf{67.71} & \textbf{21.40} & 42.59 & \textbf{61.54} & \textbf{95.70} \\
    \bottomrule
    \end{tabular}
    }
    \caption{Accuracy of \methodnames CC12M using different models for text-to-text retrieval. The LLM (Mistral) embedding is stronger than vision-language aligned embedding in terms of the unimodal text-to-text retrieval.}
    \label{tab:llm_ablation}
\end{table}

\begin{table}[ht]
\centering
\resizebox{1.\linewidth}{!}{
\begin{tabular}{@{}cccccccc@{}}
\toprule
          & (1)   & (2)   & (3)   & (4)   & (5)   & (6)   & (7)   \\ \midrule
Zeroshot  & 30.80 & 28.48 & 25.32 & 31.26 & 27.83 & \underline{32.47} & \textbf{35.81} \\
Retrieved & 26.07 & 23.75 & 27.64 & 18.00 & 13.27 & 19.76 & 22.73 \\ \bottomrule
\end{tabular}
}
\caption{Zeroshot results on Circuits with different prefixes. Numbers denote: (1) a circuit diagram of. (2) a circuit of. (3) an electronic schematic of. (4) a photo of a circuit diagram: a. (5) a picture of a \{\} circuit. (6) a photo of an electronic circuit: a. (7) a photo of a.}
\label{abl:prompting_effect}
\end{table}

\pp{Prompting strategy} We study the role of the \texttt{``{prefix}''} part of the queries when performing retrieval and when building the zero-shot weights. In Tab.~\ref{abl:prompting_effect} we show the effect of using only the zero-shot weights $\zeroshotw$ or only the retrieved captions weights $\zeroshotretrieved$ for zero-shot classification. We can see that the best prefix to build zero-shot weights is not the best prefix to retrieve information from the database, where constraining the domain in the prompt becomes fundamental to obtaining good performance. Additionally, using only retrieved captions as class prototypes leads to unsatisfactory results. We then restrict the study to two styles of the prefix: a generic \texttt{``a photo of a''} and a domain-specific \texttt{``a \{domain\} of a''}. 

We study the effect of these two prefixes in Table~\ref{abl:prompt}. For iNaturalist, since the images cover very different natural domains, \eg animals, plants, fungi, etc. we cannot design a domain-specific prompt, and we can only use the domain-agnostic one. We can observe that employing a domain-specific prefix for zero-shot and a domain-agnostic prefix for retrieval leads to generally better results across all the benchmarks.
\begin{table}[!t]
\centering
\resizebox{\linewidth}{!}{
\begin{tabular}{@{}ccccc@{}}
\toprule
\textbf{Zeroshot} & \textbf{Retrieval} & \cellcolor{CoreCircuits}\textbf{Circuits}       & \cellcolor{CoreInat}\textbf{iNaturalist} & \cellcolor{CoreHam}\textbf{HAM10000}       \\ \midrule
Domain          & Generic          & \textbf{42.94} & N/A         & \textbf{61.55} \\
Domain          & Domain           & 41.84          & N/A         & 54.40          \\
Generic         & Domain           & 41.75          & N/A         & 56.78          \\
Generic         & Generic          & 39.05          & \textbf{21.40}          & 59.30          \\ \bottomrule
\end{tabular}
}
\caption{Accuracy of our \methodnames CC12M with different prompting strategies for zero-shot weights and text-to-text retrieval. Employing a domain-specific prefix for zero-shot and a domain-agnostic prefix for retrieval leads to generally better results across all the benchmarks.}
\label{abl:prompt}
\end{table}

\section{Conclusion}

We presented \methodname, a training-free retrieval-based zero-shot solution for low-resource image classification. 
Through the retrieval of semantically relevant textual information for both image representations and class prototypes, \methodname~is able to enhance the richness of feature representations and achieves state-of-the-art performance. Remarkably, it does so without the need for additional training or labeled data, outperforming existing training-based methods under extremely low-resource conditions.
Moreover, we established a comprehensive benchmark using representative datasets and baseline models, providing a robust testing ground for the low-resource image classification task. The results demonstrate the efficacy and generalization capability of \methodname~across various low-resource domains, representing a significant step forward in low-resource image classification.

\section{Limitations}
\label{sec:limitations}

While we prove text retrieval to be a strong training-free solution on the proposed benchmark, the performance is still limited by the representation of a domain in the external database. 
An example of this low coverage is provided by~\cite{liu2023learning}, where Patch-Camelyon, a medical dataset, has a limited presence even in LAION 400M. We face the same problem when trying to apply \methodnames to Parasitic Egg Classification~\cite{anantrasirichai2022icip}, where retrieved captions from CC12M only contain the most common egg parasite name, and where supervised fine-tuning becomes the strongest way to adapt VLMs for the setting.

\section{Ethical considerations}
\label{sec:ethic}

We employ large-scale web-crawled data to enrich our representations, and this type of data is by definition mostly uncurated, and it may reflect bias from the real world. This data might contain harmful or Not Safe For Work (NSFW) content, as demonstrated in previous studies~\citep{thiel2023identifying}. 

The scientific community has acted to mitigate these risks, \eg employing NSFW filters before releasing the image-text pairs~\citep{kakaobrain2022coyo-700m}. Nevertheless, we do not employ the image content of these datasets, and we use the textual part only at the semantic level to enhance the image classification performance. Therefore we never expose users to harmful or undesired content.

\section*{Acknowledgments}

We thank CINECA and the ISCRA initiative for the availability of high-performance computing resources. This work was supported by the EU Horizon ELIAS (No. 101120237) and AI4TRUST (No.101070190) projects, and the
FAIR - Future AI Research (PE00000013), funded by NextGeneration EU, and the PRIN LEGO-AI (Prot. 2020TA3K9N) project. This work was carried out in the Vision and Learning joint laboratory of FBK and UNITN.

\bibliography{refs}

\clearpage
\appendix

\begin{table*}[ht]
\centering
\begin{adjustbox}{max width=\linewidth}
\begin{tabular}{@{}ccccccccccc@{}}
\toprule
\multirow{2}{*}{\textbf{Number of shots per class}} & \multicolumn{2}{c}{\textbf{1}} & \multicolumn{2}{c}{\textbf{5}} & \multicolumn{2}{c}{\textbf{10}} & \multicolumn{2}{c}{\textbf{15}} & \multicolumn{2}{c}{\textbf{20}} \\
& \textbf{Acc@1} & \textbf{Acc@5} & \textbf{Acc@1} & \textbf{Acc@5} & \textbf{Acc@1} & \textbf{Acc@5} & \textbf{Acc@1} & \textbf{Acc@5} & \textbf{Acc@1} & \textbf{Acc@5} \\
\midrule
ImageBind \raisebox{-1.mm}{\includegraphics[height=4.0mm]{images/icon_fire.pdf}}~\cite{zhang2024low} & 10.85&68.70 & 54.60&96.56 & 55.59&96.10 & 46.72&97.88 & 57.84&98.81 \\
SigLIP \raisebox{-1.mm}{\includegraphics[height=4.0mm]{images/icon_fire.pdf}}~\cite{zhai2023sigmoid} & 25.74&90.73 & 54.60&95.90 & 46.72&95.90 & 62.21&98.15 & 61.42&98.21  \\
\bottomrule
\end{tabular}
\end{adjustbox}
\caption{Accuracy of training-based \raisebox{-1.mm}{\includegraphics[height=4.0mm]{images/icon_fire.pdf}} solutions with increasing number of training shots per class on HAM10000.}
\label{abl:shots}
\end{table*}

\section{Baselines implementation details}
\label{app:implementation}

For VLMs (\ie CLIP, ImageBind, and SigLIP) we report the zero-shot results with zero-shot weights built starting from the prompt \texttt{``\{prefix\}~[CLS]"}. As \texttt{\{prefix\}} we test both domain-specific (\eg \texttt{``a circuit diagram of a"} for Circuits dataset) and domain-agnostic sentences (\texttt{``a photo of a"} as standard practice in CLIP), and we report the best result for each model. For iNaturalist we follow the insights of~\cite{parashar2023prompting} and try using common names of animals and plants instead of their scientific names. We find that merging the zero-shot weights of common and scientific names improves the overall accuracy, therefore we report these results for this dataset. 

For LMM (\ie BLIP2 and LLaVA) we feed the query image with the textual prompt \texttt{``Question: what's the name of the object in the image out of [class names]? Answer with the name only. Answer: "}, then parse the answer and match the name with the dataset classes.

For ImageBind\raisebox{-1.mm}{\includegraphics[height=4.0mm]{images/icon_fire.pdf}} and SigLIP\raisebox{-1.mm}{\includegraphics[height=4.0mm]{images/icon_fire.pdf}} we follow~\citep{zhang2024low} recipe for fine-tuning, and generate synthetic images using their pipeline. The generation involves using a Stable Diffusion model~\citep{rombach2021high} on top of noised images, and feeding the model with a domain-specific prompt, \eg \texttt{``a circuit diagram of a [CLS]."} to re-generate the missing part. The amount of noise added from the diffusion schedule depends on the role of the synthetic image, \ie it is set to 30\% of the schedule for samples used as positive in the contrastive loss, and 60\% for the negative samples. %
We maintain the original training hyperparameters (including batch size and learning rate), and we train an AdaptFormer~\cite{chen2022adaptformer} module with rank 2 and a linear classification head.

\section{On synthetic data augmentation}
\label{app:synth}

We extend the discussion from Sec.~\ref{sec:intro} on using synthetic data augmentation to address low-data scenarios. We argue that while synthetic data generation can effectively enhance recognition performance, it introduces training images that are either overly similar to the original samples or incorrect, violating domain rules, one example being the case of~\cite{zhang2024low}. We showcase examples of both ``positive'' and ``negative'' types of synthetic augmentation in Fig.~\ref{fig:synthetic}. We can see that ``positive'' samples do not differ significantly from the original sample and do not bring meaningful variation to the original sample. On the other hand, the ``negative'' samples differ to the point of changing semantics (first row) or breaking the reference domain rules (third row).

\begin{table*}[th]
    \setlength{\tabcolsep}{0.5pt}
\renewcommand{\arraystretch}{1.}
    \centering
    \begin{tabular}{ccc|ccc}
    \toprule
        \textbf{Original} & \textbf{Positive} & \textbf{Negative} & \textbf{Original} & \textbf{Positive} & \textbf{Negative} \\
        \midrule

        \raisebox{+.15\height}{\includegraphics[width=2.5cm]{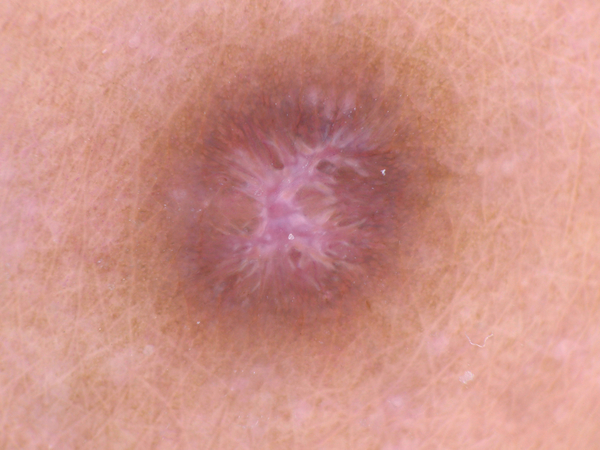}} & \includegraphics[width=2.5cm]{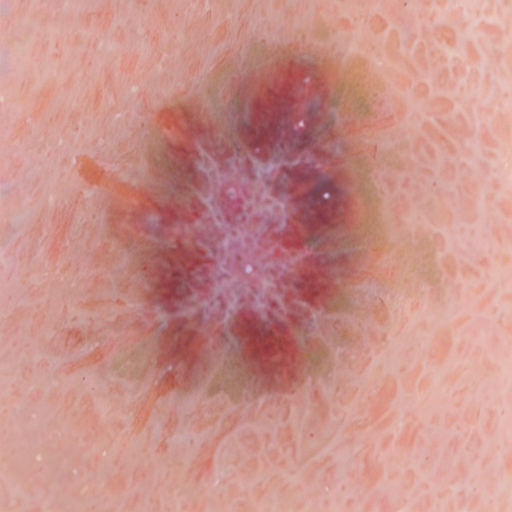} & \includegraphics[width=2.5cm]{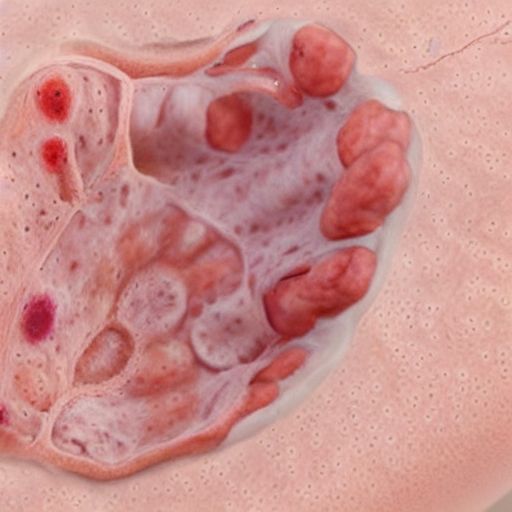} & \raisebox{+.15\height}{\includegraphics[width=2.5cm]{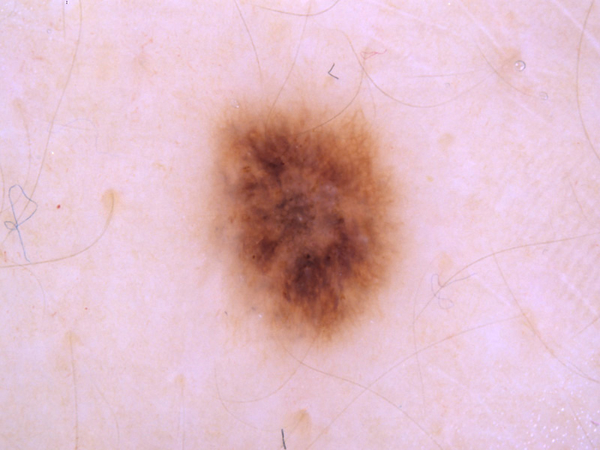}} & \includegraphics[width=2.5cm]{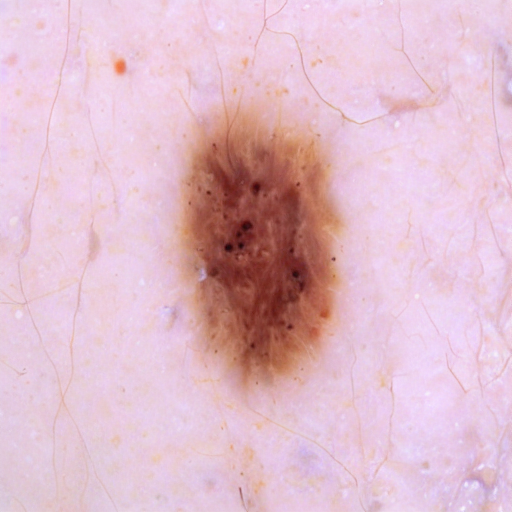} & \includegraphics[width=2.5cm]{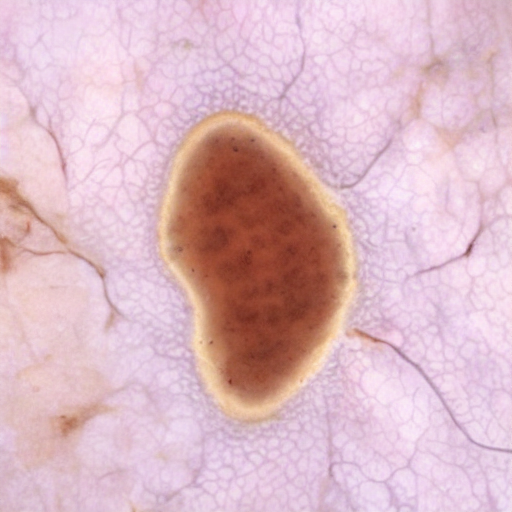} \\

        \raisebox{+.15\height}{\includegraphics[width=2.5cm]{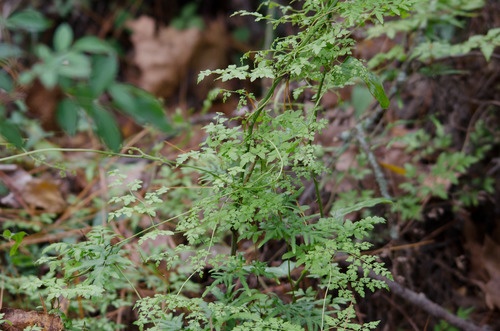}} & \includegraphics[width=2.5cm]{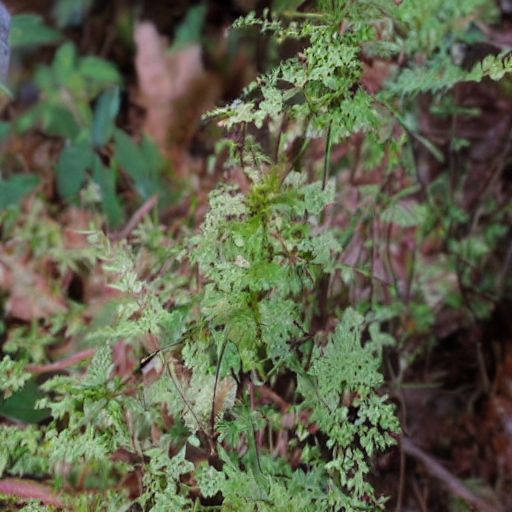} & \includegraphics[width=2.5cm]{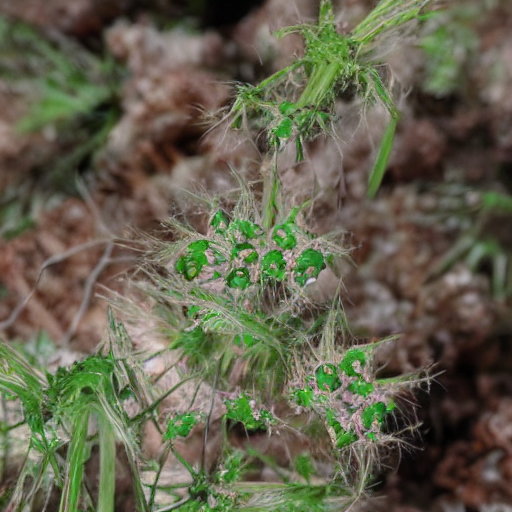} &

        \raisebox{-.15\height}{\includegraphics[width=2.5cm]{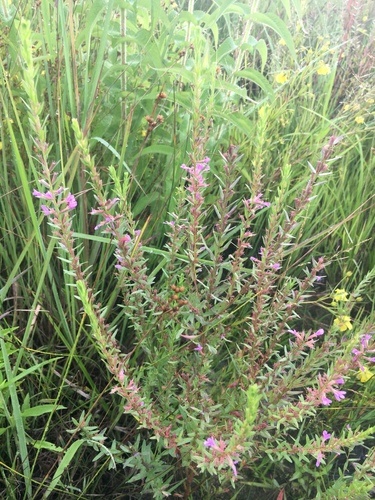}} & \includegraphics[width=2.5cm]{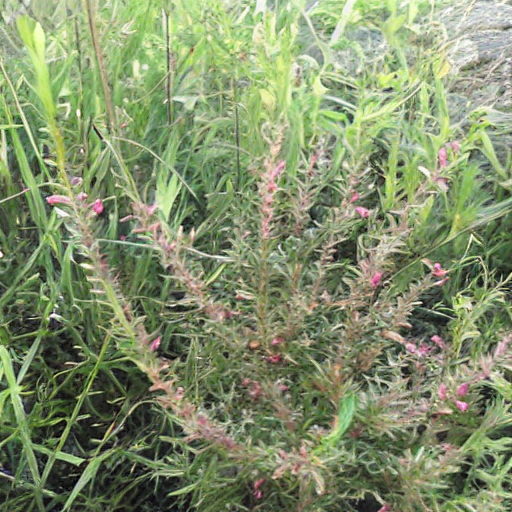} & \includegraphics[width=2.5cm]{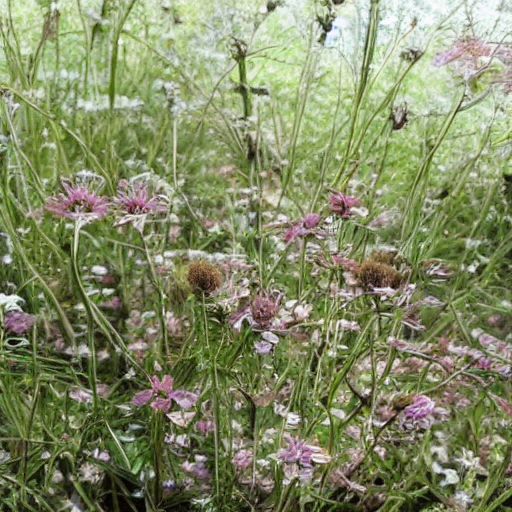} \\

        \raisebox{+.45\height}{\includegraphics[width=2.5cm]{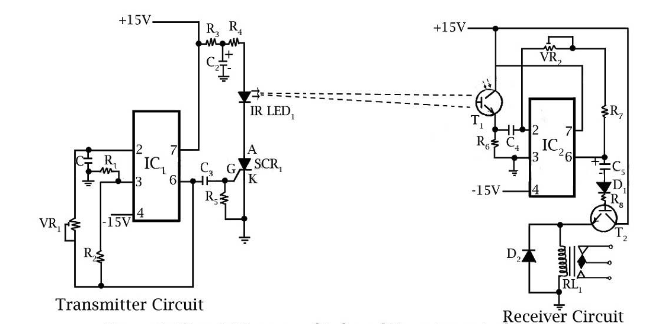}} & \includegraphics[width=2.5cm]{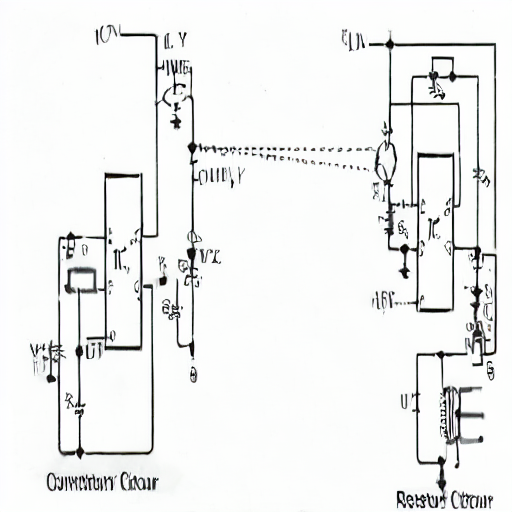} & \includegraphics[width=2.5cm]{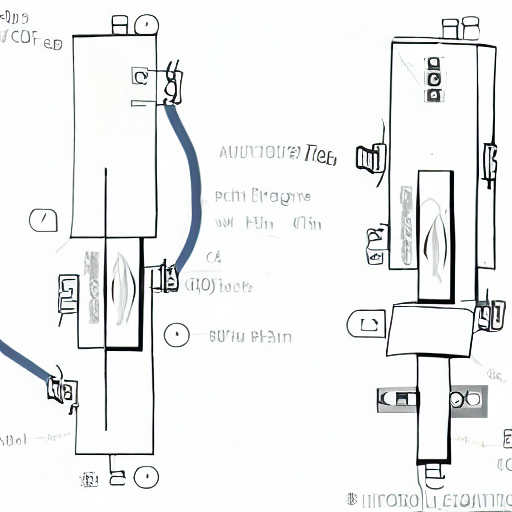} &

        \raisebox{+.1\height}{\includegraphics[width=2.5cm]{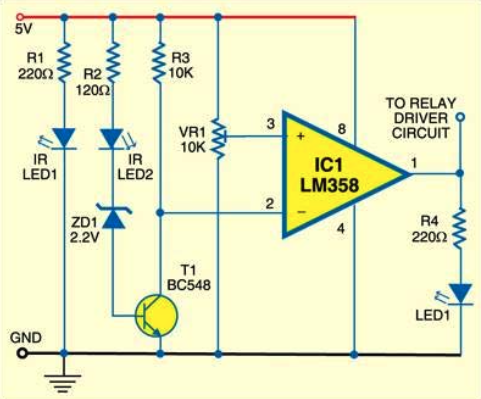}} & \includegraphics[width=2.5cm]{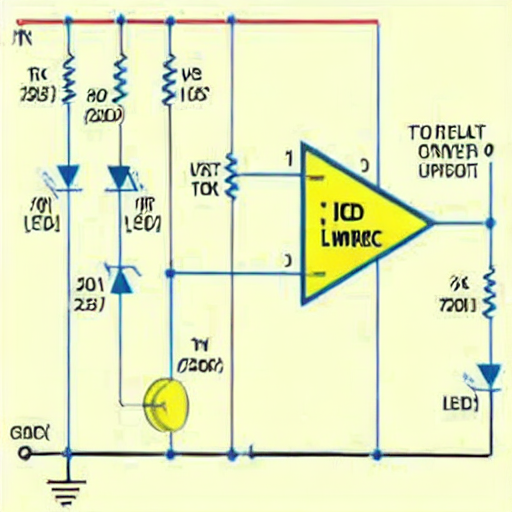} & \includegraphics[width=2.5cm]{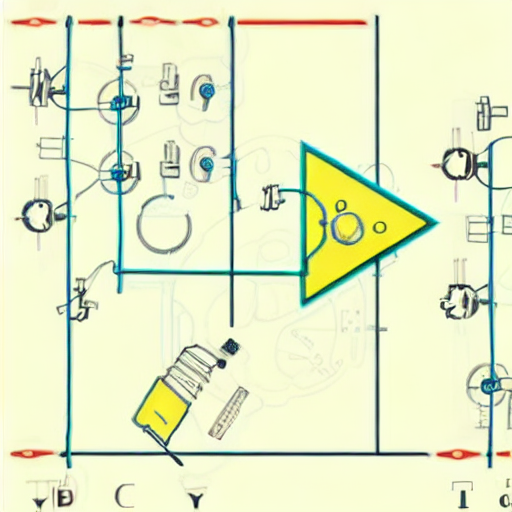} \\

        \bottomrule
    \end{tabular}
    \caption{Synthetic image from the baseline of~\citep{zhang2024low}. We show original samples, the ``positive'' augmentation and the ``negative'' augmentation.}
    \label{fig:synthetic}
\end{table*}

\section{On having an image-to-image retrieval}

\begin{table}[!ht]
    \centering
\resizebox{\linewidth}{!}{
    \begin{tabular}{cccc}
    \toprule
       \textbf{Method}  & \cellcolor{CoreCircuits}\textbf{Circuits} & \cellcolor{CoreInat}\textbf{iNaturalist} & \cellcolor{CoreHam}\textbf{HAM10000} \\
       \midrule

       DINOv2 img2img & 23.28 & 7.78 & 59.96 \\

       \methodnames (Ours --- CC12M) & 42.94 & 21.40 & 61.54 \\
       \bottomrule
    \end{tabular}
    }
    \caption{Top-1 accuracy on the proposed benchmark using DINOv2 features to retrieve relevant images compared to our \methodname CC12M.}
    \label{tab:dino_retrieval}
\end{table}

Initially, we also implemented an image-to-image retrieval system leveraging DINOv2 embeddings. We hypothesized that enhancing the original image embedding with features from retrieved images would improve its capability to retrieve relevant textual information. However, this approach did not yield the desired results. Although the retrieved images were from the correct domain, they often lacked the appropriate semantic class. This mismatch caused the embeddings to deviate significantly from the target class, resulting in a text-retrieval performance decline. Quantitative results supporting this observation are presented in Tab.~\ref{tab:dino_retrieval}.

\section{Interesting failure cases}

We extend the discussion started in Section~\ref{sec:limitations} regarding the failure cases of retrieval-based solutions \wrt to training-based ones. We include the quantitative results on the Parasitic Egg Recognition challenge~\cite{anantrasirichai2022icip} in Tab.~\ref{tab:egg_parasites}. We can see that in this case, the training-based solutions outperform the training-free solutions by a large margin. We deem this for two reasons: i) the classes differ visually, making a training approach powerful in telling the different classes, and ii) this type of data is under-represented in large-scale datasets~\cite{liu2023learning}, making the retrieval less effective which, nevertheless, outperforms the other training-free solutions.

\begin{table}[ht]
    \centering
    \begin{tabular}{cc}
    \toprule
    \textbf{Method} & \textbf{Acc@1} \\ \midrule

    ImageBind \raisebox{-1.mm}{\includegraphics[height=4.0mm]{images/icon_fire.pdf}}~\cite{zhang2024low} & 33.27 \\
     SigLIP \raisebox{-1.mm}{\includegraphics[height=4.0mm]{images/icon_fire.pdf}}~\cite{zhai2023sigmoid} & 17.59 \\ \midrule
     ImageBind~\cite{girdhar2023imagebind}    & 9.18 \\
     SigLIP~\cite{zhai2023sigmoid}    & 12.72 \\
     \methodnames (Ours --- CC12M) & 15.14 \\
      \bottomrule
    \end{tabular}
    \caption{Top-1 accuracy on the Parasitic Egg Classification.}
    \label{tab:egg_parasites}
\end{table}

\section{Additional analyses}

\pp{Dependency on sample quantity}
The experiments in Tab.~\ref{tab:quantitative} are conducted with $\sim5$ annotated samples per class as in the Circuits dataset of~\citep{zhang2024low}, where only 154 images are available for training on 32 classes. We complement this analysis by showing in Tab.~\ref{abl:shots} the effect of having access to more training data for training-based solutions on HAM10000. 

We can see that the model performance scales with the amount of annotated samples per class. SigLIP trained with 15 samples achieves better results than \methodname. The low amount of training and validation data, on the other hand, makes the selection of the best-trained model noisy, therefore the improvement with more samples is not straightforward to assess. 

\section{Accuracy@1 vs Accuracy@5 tradeoff}

\begin{table}[!t]
\centering
\resizebox{0.5\columnwidth}{!}{
\begin{tabular}{cc}
\toprule
Acc@1 & Acc@5 \\
\midrule
27.73 & 96.10 \\
39.58 & 95.96 \\
54.80 & 95.63 \\
57.97 & 95.50 \\
61.95 & 95.30 \\
62.21 & 94.51 \\
\bottomrule
\end{tabular}}
\caption{Acc@1 vs Acc@5 tradeoff on HAM10000.}
\label{tab:pareto_frontier}
\end{table}

In our investigation, we observed that modifying the hyperparameters $\alpha$, $\beta$, and $\tau$ resulted in a trade-off between Accuracy@1 and Accuracy@5. We prioritized Accuracy@1, which is widely regarded as the primary classification metric. This sensitivity was particularly pronounced in the HAM10000 dataset, where we identified several data points that constitute the Pareto frontier when plotting Accuracy@1 against Accuracy@5. Some notable examples from this frontier are presented in Table~\ref{tab:pareto_frontier}.

\section{Retrieval Efficiency}

Our method \methodnames leverages an efficient indexing and retrieval mechanism, implemented with FAISS~\cite{douze2024faiss}, achieving 57.54 ms per image-text retrieval on a CPU. As text-to-text retrieval is performed offline, we do not include it in the runtime analysis. As shown in Table~\ref{tab:retrieval_efficiency}, per image inference at runtime, \methodnames requires on average 110 ms (the sum of SigLIP inference time and retrieval time), while the competitor with the best performance (SigLIP@384px) requires 53 ms. \methodnames requires 110 ms, which is about one-third of other VLMs (e.g. CLIP ViT-L@336px, BLIP2-EVA) and is of one magnitude less than LLaVA 1.6 (34B). 

In terms of memory/storage, in addition to the models, \methodnames requires the storage of both the metadata (i.e., the filenames and original captions), and the indexes (i.e., the embeddings). For CC12M, the metadata is of 2.7 GB and the index is of 3.4 GB by SigLIP (3.1 GB by SFR-Embedding-Mistral). For our COYO-700M subset, the metadata is of 6.0 GB, and the indices are of 22 GB by SigLIP (23 GB by SFR-Embedding-Mistral). In comparison, other competitors do not require the storage of a database, but the models themselves could be storage-demanding, e.g. LLaVA 1.6 (34B) occupies about 65GB of storage, while CoRE (COYO-700M) in total occupies as little as 9.4GB, as the LLM is optional as shown in the ablation ``On the use of LLM for embedding''. 

\begin{table*}[!th]
    \centering
    \resizebox{\linewidth}{!}{
    \begin{tabular}{cccc}
    \toprule
        Method & Inference (ms) & Params & Storage (GB) \\
        \midrule
        CLIP ViT-L & 55 & 123M $f_{VLM}^{T}$ + 303M $f_{VLM}^{I}$ & 1.6  \\
        CLIP ViT-L@336px & 368 & 123M $f_{VLM}^{T}$ + 304M $f_{VLM}^{I}$ & 1.6 \\
        BLIP2-EVA & 397 & 2.9B $f_{LLM}$ + 986M $f_{VLM}^{I}$ & 2.3 \\
        LLaVA 1.6 (34B) & 4970 & 34B $f_{LLM}$ + 303M $f_{VLM}^{I}$ & 65 \\
        ImageBind & 52 & 302M $f_{VLM}^{T}$ + 632M $f_{VLM}^{I}$ & 3.7 \\
        SigLIP@384px & 53 & 449M $f_{VLM}^{T}$ + 428M $f_{VLM}^{I}$ & 3.3 \\
        \midrule
        \textsc{CoRE} & 110 & 449M $f_{VLM}^{T}$ + 428M $f_{VLM}^{I}$ + 7B $f_{LLM}$* & 17.3 + 6.1 (CC12M) \\
        \bottomrule
    \end{tabular}}
    \caption{Retrieval efficiency of our \methodnames and all the baselines in terms of inference time, parameter count, and storage requirements.}
    \label{tab:retrieval_efficiency}
\end{table*}

\section{Examples of retrieved captions}

We provide a qualitative showcase of retrieved captions by out \methodnames for both text-to-text and image-to-text in Tab.~\ref{tab:retrieval_samples_txt} and Tab.~\ref{tab:retrieval_samples_img}. We use the 10\% subset of COYO-700M as the database, and we also show some failure cases (\eg sixth row of Tab.~\ref{tab:retrieval_samples_img}).

\begin{table*}[th]
    \centering
    \renewcommand{\arraystretch}{3}
    \resizebox{\linewidth}{!}{
    \begin{tabular}{c|ccc}
    \toprule
    \textbf{Input} & \multicolumn{3}{c}{\textbf{Retrieved captions}} \\ \midrule
      \makecell{A circuit diagram of \\an amplifier.} & \makecell{\makecell{Electrical Diagram\\Amplifier}} & \makecell{Audio amplifier circuit\\diagram with layout} & \makecell{Circuit diagram for transistor\\as audio amplifier} \\

      \makecell{A circuit diagram\\of a LED.} & \makecell{\makecell{Technical Drawing\\of an LED}} & \makecell{The symbol for a \\light emitting diode.} & \makecell{One LED with leads\\ all bent out} \\

      \makecell{A circuit diagram of \\an audio mixer.} & \makecell{\makecell{Photo of an \\audio mixer board}} & \makecell{8-Channel Audio\\Mixer picture 1} & \makecell{Mixer with volume\\ faders and pan knobs} \\ \midrule

      \makecell{A skin lesion of \\Bowen's disease.} & \makecell{\makecell{Pre-Cancerous\\Actinic Keratosis}} & \makecell{A biopsy specimen \\showing hyperkeratosis, papillom} & \makecell{Looking for\\premalignant skin cancer} \\ 

      \makecell{A skin lesion\\of melanoma.} & \makecell{\makecell{Melanoma of \\the Skin, Cut-section}} & \makecell{This picture shows\\a melanoma lesion\\with varying colors.} & \makecell{A mole that\\turned out to be\\melanoma skin cancer} \\ 

      \makecell{A skin lesion of \\melanocytic nevi.} & \makecell{\makecell{This picture shows\\a melanoma lesion\\with varying colors.}} & \makecell{Recurrence of\\ melanocytic naevus} & \makecell{Graphic of\\a melanoma} \\ \midrule

      \makecell{A photo of a\\Lygodium japonicum.} & \makecell{\makecell{A picture of a\\Japanese holly fern}} & \makecell{japanese painted fern \\has silver metallic fronds} & \makecell{Pyrrosia Ferns on\\ moss covered trunk} \\

      \makecell{A photo of a\\Salvinia minima.} & \makecell{\makecell{12 Water Spangles(Salvinia\\Minima), Live Aquarium/Aquatic}} & \makecell{Salvinia Minima (Water\\Spangles) floating aquarium plant} & \makecell{water drop salvinia \\sp trichomes stock image} \\

      \makecell{A photo of a\\Azolla filiculoides.} & \makecell{\makecell{The Azolla fern\\has leaves floating\\on the water surface}} & \makecell{Image of Azolla \\filiculoides(). Click\\to enlarge parts of image.} & \makecell{Picture of Fern} \\
      
    \bottomrule
    \end{tabular}
    }
    \caption{Examples of retrieved captions in text-to-text using the COYO-700M subset.}
    \label{tab:retrieval_samples_txt}
\end{table*}

\begin{table*}[th]
    \centering
    \renewcommand{\arraystretch}{3}
    \resizebox{\linewidth}{!}{
    \begin{tabular}{c|ccc}
    \toprule
    \textbf{Input} & \multicolumn{3}{c}{\textbf{Retrieved captions}} \\ \midrule
      \includegraphics[width=4.5cm]{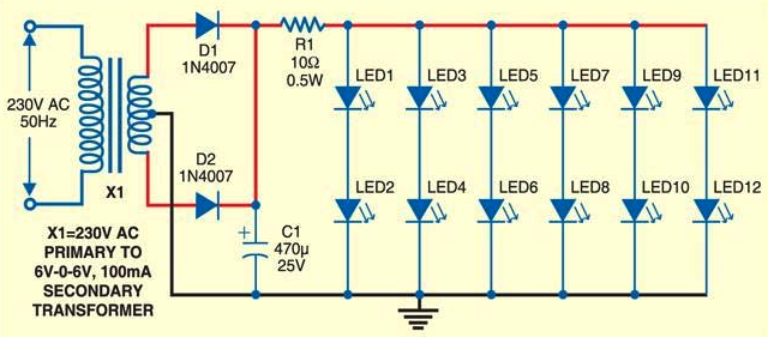} & \makecell[bc]{Hobby Electronics\\Circuits: AC Powered 220V\\Led Circuit} & \makecell[bc]{Transformerless Led\\ Lighting Led Lamp\\Circuit Electronics} & \makecell[bc]{Astonishing\\Christmas Lights} \\

      \includegraphics[width=4.5cm]{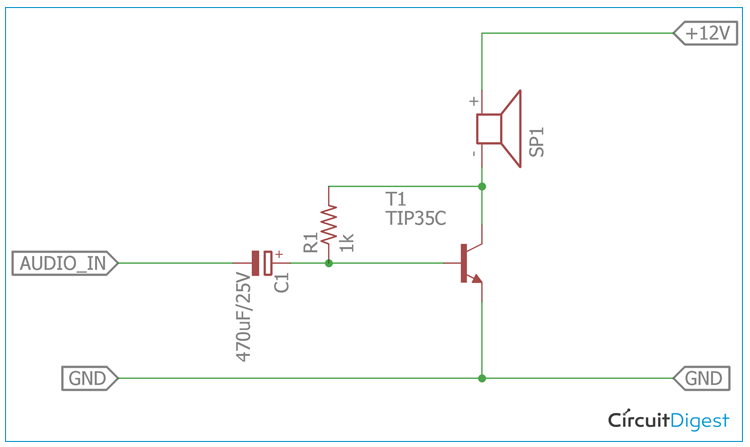} & \makecell[bc]{PIEZO SOUNDER\\WITH BUILT-IN CIRCUIT} & \makecell[bc]{How To Build\\A Speaker Circuit\\With Adjustable Volume} & \makecell[bc]{What all do I\\need for a simple\\speaker circuit?} \\

      \includegraphics[width=3.5cm]{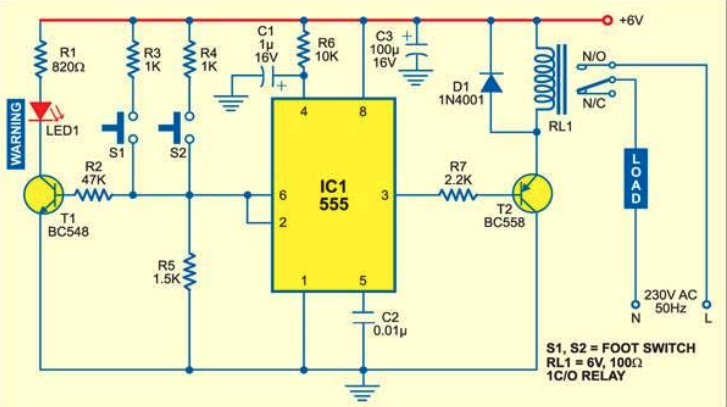} & \makecell[bc]{IC 555 dry\\run protection} & \makecell[bc]{Wireless Remote \\Control Switch} & \makecell[bc]{230V AC Mains Over\\ Voltage Protection Circuit} \\ \midrule

      \includegraphics[width=3.5cm]{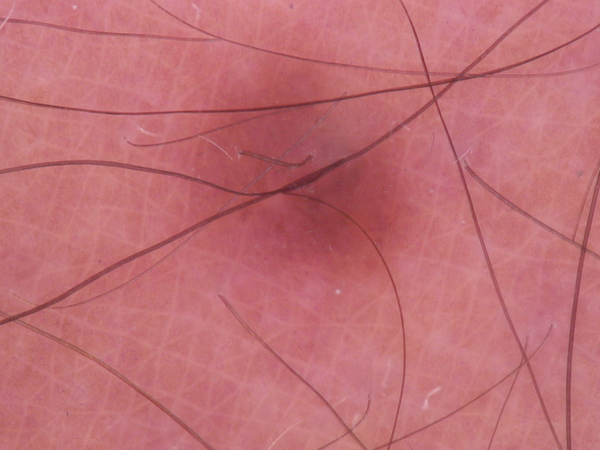} & \makecell[bc]{Skin coloured papules\\ centred around\\hair follicles.} & \makecell[bc]{Dermoscopic image of\\ a porokeratosis of Mibelli
 lesion} & \makecell[bc]{Pigmented basal cell\\carcinoma dermoscopy} \\ 

      \includegraphics[width=3.5cm]{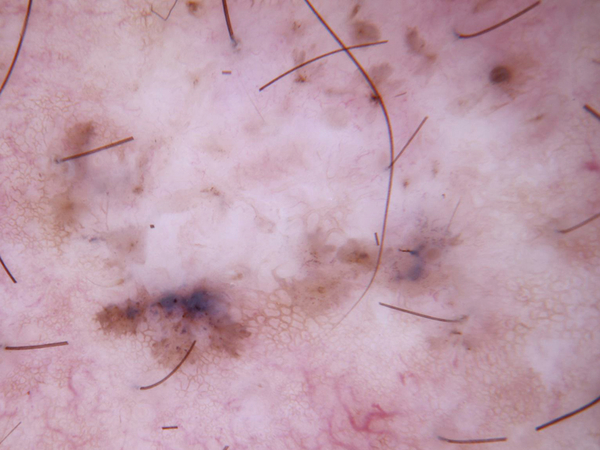} & \makecell[bc]{Dermoscopy. Brown\\and blue-grey dots/clods.} & \makecell[bc]{Dermoscopic image of a\\porokeratosis of Mibelli
lesion} & \makecell[bc]{Dermoscopy. Chaos \\and clues} \\ 

      \includegraphics[width=4.5cm]{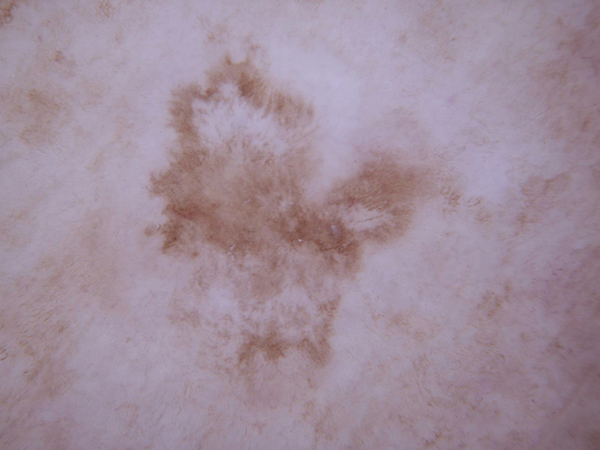} & \makecell[bc]{brown blotches
\\are formed} & \makecell[bc]{Oil Red Staining.} & \makecell[bc]{pigment used as\\normal pigment pattern} \\ \midrule

      \includegraphics[width=2.0cm]{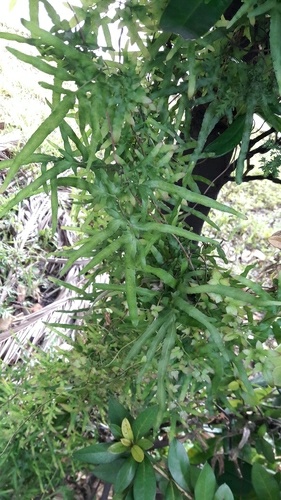} & \makecell[bc]{Notogrammitis billardierei \\(Finger fern) at Wingecarribee} & \makecell[bc]{Asplenium polyodon (East Maui)\\This image is licensed} & \makecell[bc]{Not sure what this fern is\\I thought maybe Buckler\\fern. Any ideas?} \\

      \includegraphics[width=3.5cm]{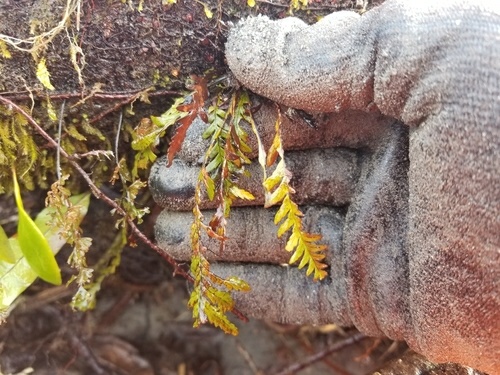} & \makecell[bc]{A seed fern f
rond\\is prepared for analysis.} & \makecell[bc]{A tiny plant on\\a tree fern's trunk} & \makecell[bc]{Asplenium polypodon\\(West Maui)} \\

      \includegraphics[width=2.0cm]{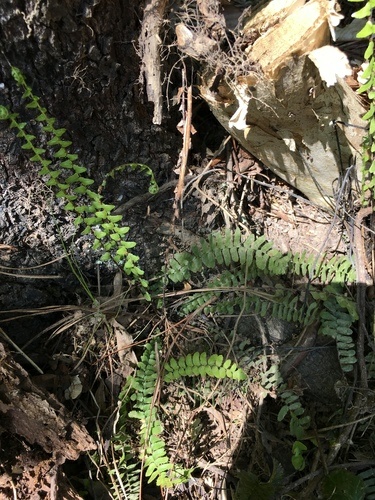} & \makecell[bc]{Astrolepis cochisensis\\Cochise Scaly Cloakfern} & \makecell[bc]{Notogrammitis billardierei\\(Finger fern)} & \makecell[bc]{Ferns emerging\\from charred earth} \\
      
    \bottomrule
    \end{tabular}
    }
    \caption{Examples of retrieved captions in image-to-text using the COYO-700M subset.}
    \label{tab:retrieval_samples_img}
\end{table*}

\section{Computational requirements}

Synthetic data generation for trained baselines, as the amount of training data is low, has an upper bound of 12 GPU/hours on an NVIDIA A100 for iNaturalist. The subsequent model training has an upper bound of 4 GPU/hours on an NVIDIA A100 for SigLIP.

For our \methodname, the most time-consuming task is represented by the external database embedding. The upper bound is 16 GPU/days for the COYO-700M subset. CC12M has been embedded in 3 GPU/days. Retrieval can then be performed without access to any GPU and on any consumer hardware in $\sim 58$ms.

\section{Licenses}

Most of the datasets used (Circuits~\cite{zhang2024low}, iNaturalist~\cite{van2017inaturalist}, HAM10000~\cite{tschandl2018ham10000}, and Parasitic Egg Classification~\cite{anantrasirichai2022icip}) are released under Creative Commons Attribution Non-Commercial 4.0. COYO-700M~\cite{kakaobrain2022coyo-700m} is released under Creative Commons Attribution 4.0, while CC12M~\cite{changpinyo2021cc12m} is released ``as is''.

LLaVA~\cite{liu2023improved} and SigLIP~\cite{zhai2023sigmoid} are released under Apache 2.0. ImageBind~\cite{girdhar2023imagebind} is released under Creative Commons Attribution Non-Commercial ShareAlike 4.0. CLIP~\cite{radford2021learning} is released under MIT. BLIP~\cite{li2023blip} is released under BSD-3-Clause. SFR-Embedding-Mistral~\cite{SFRAIResearch2024} is released under Creative Commons Attribution Non-Commercial 4.0.

PyTorch~\cite{Ansel_PyTorch_2_Faster_2024} is released ``as is''. Faiss~\cite{douze2024faiss} is released under MIT.

\end{document}